\documentclass[conference]{IEEEtran}
\usepackage[latin9]{inputenc}
\usepackage{amsmath}
\usepackage{graphicx}

\makeatletter

\def\ps@IEEEtitlepagestyle{
  \def\@oddfoot{\mycopyrightnotice}
  \def\@evenfoot{}
}
\def\mycopyrightnotice{
  {\footnotesize 979-8-3503-5028-9/24/\$31.00~\copyright~2024 IEEE\hfill} 
  \gdef\mycopyrightnotice{}
}

\@ifundefined{showcaptionsetup}{}{
 \PassOptionsToPackage{caption=false}{subfig}}
\usepackage{subfig}
\makeatother

\usepackage{eso-pic}
\newcommand\AtPageUpperMyright[1]{\AtPageUpperLeft{
 \put(\LenToUnit{1cm},\LenToUnit{-1cm}){
     \parbox{0.6\textwidth}{\raggedright\fontsize{9}{11}\selectfont #1}}
 }}
\newcommand{\conf}[1]{
\AddToShipoutPictureBG*{
\AtPageUpperMyright{#1}
}
}

\begin{document}

\title{A Machine Learning Approach for Crop Yield and Disease Prediction Integrating Soil Nutrition and Weather Factors}
\conf{2024 International Conference on Advances in Computing, Communication, Electrical, and Smart Systems (iCACCESS), 8-9 March, Dhaka, Bangladesh} 

\author{\IEEEauthorblockN{\small Forkan Uddin Ahmed}
\IEEEauthorblockA{\small Department of CSE\\
Chittagong University of\\
Engineering \& Technology\\
Chattogram, Bangladesh\\
forkan1510699@gmail.com}
\and
\IEEEauthorblockN{\small Annesha Das}
\IEEEauthorblockA{\small Department of CSE\\
Chittagong University of\\
Engineering \& Technology\\
Chattogram, Bangladesh\\
annesha@cuet.ac.bd}
\and
\IEEEauthorblockN{\small Md. Zubair}
\IEEEauthorblockA{\small Department of CSE\\
Chittagong University of\\
Engineering \& Technology\\
Chattogram, Bangladesh\\
zubairhossain773@gmail.com}}
\maketitle

\begin{abstract}
The development of an intelligent agricultural decision-supporting system for crop selection and disease forecasting in Bangladesh is the main objective of this work. The economy of the nation depends heavily on agriculture. However, choosing crops with better production rates and efficiently controlling crop disease are obstacles that farmers have to face. These issues are addressed in this research by utilizing machine learning methods and real-world datasets. The recommended approach uses a variety of datasets on the production of crops, soil conditions, agro-meteorological regions, crop disease, and meteorological factors. These datasets offer insightful information on disease trends, soil nutrition demand of crops, and agricultural production history. By incorporating this knowledge, the model first recommends the list of primarily selected crops based on the soil nutrition of a particular user location. Then the predictions of meteorological variables like temperature, rainfall, and humidity are made using SARIMAX models. These weather predictions are then used to forecast the possibilities of diseases for the primary crops list by utilizing the support vector classifier. Finally, the developed model makes use of the decision tree regression model to forecast crop yield and provides a final crop list along with associated possible disease forecast. Utilizing the outcome of the model, farmers may choose the best productive crops as well as prevent crop diseases and reduce output losses by taking preventive actions. Consequently, planning and decision-making processes are supported and farmers can predict possible crop yields. Overall, by offering a detailed decision support system for crop selection and disease prediction, this work can play a vital role in advancing agricultural practices in Bangladesh. 
\end{abstract}

\begin{IEEEkeywords}
Agriculture, Crop suggesting model, SARIMAX, Crop production, Crop disease, Haversine formula, Machine learning.
\end{IEEEkeywords}

\section{Introduction}
Agriculture remains a central lifeline of Bangladesh's economy, contributing to 12\% of the nation's GDP and employing 45\% of its workforce \cite{1}. Nevertheless, increased population pressure and various challenges are testing the sector's capabilities. Key problems include limited knowledge of appropriate crop selection considering soil nutrition, weather forecasting limitations, and vulnerability to pests and diseases. Modernizing agriculture is essential, and the implementation of machine learning and artificial intelligence knowledge in the age of the fourth industrial revolution can instigate such transformation.

Crops cultivated in Bangladesh are influenced by numerous factors including soil nutrients, weather, and disease risks, all varying across the country's regions. Recognizing these regulators is crucial as not all crops are suitable for all areas. Soil testing is particularly important for understanding soil composition and nutrient levels, paving the way for better crop selection \cite{2}. Machine learning can streamline this process by analyzing soil states and suggesting suitable crops accordingly.

Weather forecasting is another crucial determinant of crop production \cite{3}. By incorporating machine learning and AI into weather prediction, more accurate forecasts can be generated to support crop selection. Furthermore, machine learning's predictive capabilities can help to assess the risk level of crop diseases based on weather parameters such as temperature, rainfall, humidity and, thereby influencing crop choices.

Despite significant research in crop suggestions based on various parameters, no work has yet combined weather forecasts, soil nutrition, and disease prediction for improved crop productivity in Bangladesh. The primary objective is to blend these factors for crop recommendation. The main challenge lies in merging disease prediction with weather parameters to enhance the quality of suggestions. Data related to disease risks and weather conditions need to be meticulously organized and categorized for different crops.

The contributions of the proposed work can be summarized as follows:

\begin{itemize}
\item A unified framework is created for weather prediction in Bangladesh, integrating different meteorological variables using the SARIMAX time series model.

\item A dataset of the diseases of different crops based on weather parameters (temperature and humidity) is developed.

\item A user-friendly crop-suggesting model is introduced that aids farmers in making efficient decisions based on their soil attributes, weather forecasts, and disease risks, thereby enhancing their crop yield and profitability.
\end{itemize}
Section II delves into a review of existing literature on this topic. Our methodology is depicted thoroughly in Section III. A practical assessment of our model is exhibited in Section IV, and the paper is concluded in Section V.

\section{Related Work}
Several researchers have been attempting to use modern approaches in the agricultural field to improve crop production recently.
Hatfield et al. \cite{5} used the Seasonal Autoregressive Integrated Moving Average model, and for estimating the crop's yield, they utilized the random forest regression technique. However, they did not include the impact of diseases on crop production. 
S. Khaki et al. \cite{6} introduced a deep neural network model for forecasting crop production. They identified the significant influence of weather factors on crop production compared to genotype. They also employed the neural network model to predict the weather.
Saranya et al. \cite{2} used SVM, KNN, and logistic regression to predict the best crop according to soil tests and weather forecasts. Elavarasan et al. \cite{7} applied various machine learning models, both supervised and unsupervised, to
guess crop production and saw that the expectation-maximization algorithm and the support vector machine gave finer results than the other algorithms based on various error measurement approaches. Khattab et al. \cite{8} focused on disease prediction
based on weather parameters. They made use of an IoT-based monitoring system to forecast early plant disease. Ryan et al. \cite{9} utilised Markov random fields, which are responsible for the spatial element among neighbouring sites for herbicide resistance. Afrin et al. \cite{10} focused on crop yield prediction by analyzing soil properties of 28 sub-districts of Bangladesh. Crop yield predictions were made by using DBSCAN, PAM, CLARA, K-means, and other data mining techniques and four different types of linear regression. Parveen et al. \cite{11} gave a review of machine learning methods for agricultural crop disease prediction. It discussed the integration of meteorological factors and historical disease data and stressed the potential of machine learning in enhancing disease management tactics. Limitations in data availability and quality were found, particularly in emerging areas. The intricacy of disease connections and the requirement for precise and timely disease data were also stressed. An early crop disease prediction method using machine learning was described by Vijayakumar et al. \cite{12}. It investigated the use of meteorological variables and physiological information about plants to create forecasting models. To safeguard crops, the study emphasized the value of early disease identification. The complexity of disease interactions and the requirement for real-time data updates were also cited in the article as major obstacles to creating reliable and accurate disease prediction models.
Ahmed et al. \cite{13} used a Human-Computer Interaction (HCI) oriented
method namely Soft System Methodology (SSM) along with different machine learning models such as Naive Bayes, j48, Sequential Minimal Optimization, and Multiclass classifier for predicting crop yields. Aggarwal et al. \cite{14} proposed a machine learning-based integrated solution for crop recommendation and yield prediction. It developed thorough models by combining soil properties, meteorological variables, and historical yield data. The study highlighted the advantages of including these aspects while making agricultural decisions. The report discussed the difficulties in gathering data, particularly when it comes to precise and dependable soil and meteorological data. It also suggested that models needed to be improved to fit certain agroecological zones.

\section{Outline of Methodology}
\begin{figure}[htbp]
    \centering
    \includegraphics [scale=0.58]{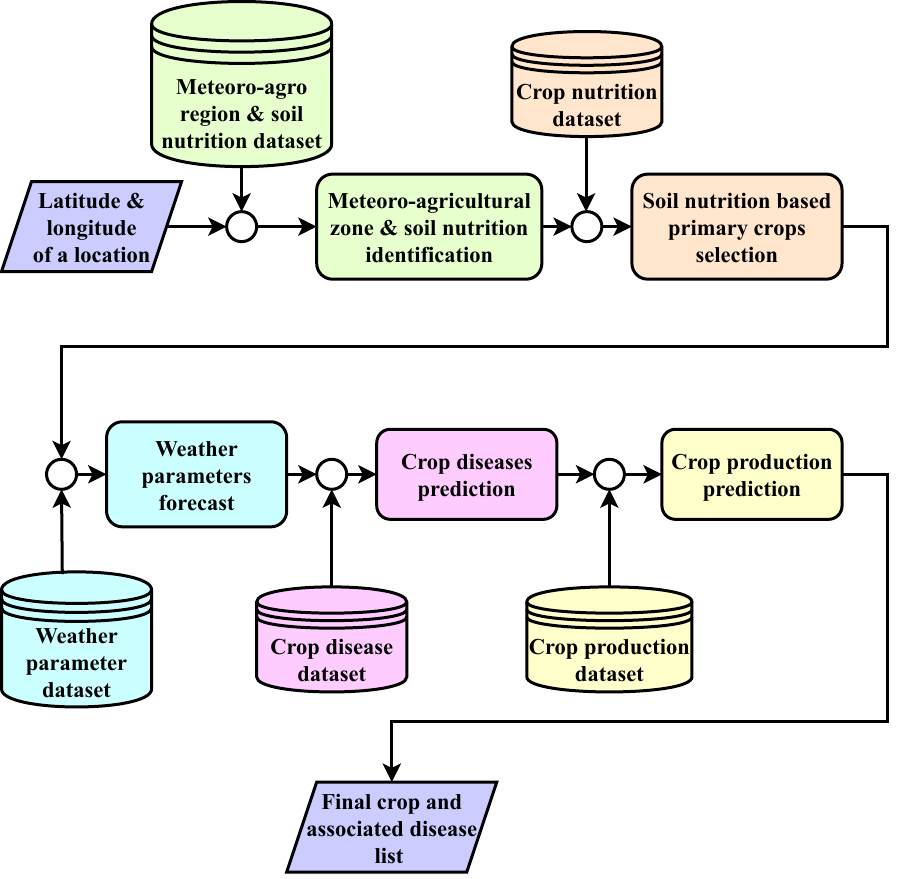}
    \caption{Overall system architecture }
    \label{fig: Overall_System_Architecture.drawio}
\end{figure}

This section describes the suggested approach for our work. Bangladesh is mainly an agricultural country and its economic development vastly depends on the advancement of the agricultural sector. But this field faces a lot of difficulties. The cultivable land is reducing over the years when the demand for production is increasing. climate change and its consequences are also making a large impact on the production of agricultural products. It was challenging to identify the most productive crops. We have to follow a sequential process to get short-listed suggestions for the top-producing crops and the possible disease vulnerability forecast for each crop on the suggested list. Making real-world datasets is challenging because crop productivity is heavily influenced by
weather, soil quality, season, and disease possibility. The methodology contains
several steps, depicted in Fig.~\ref{fig: Overall_System_Architecture.drawio}. The steps applied in the methodology are discussed in the next sections.

\subsection{Dataset Preparation}
Bangladesh has 64 districts, 492 sub-districts, and 30 agroecological zones \cite{13}. We have used seven different datasets for the proposed work. Details
about each of the datasets are presented in the following Table I and discussed in the following sections.

\begin{table}[htbp]
    \centering
    \caption{DATASET OVERVIEW}
    \begin{tabular}{|p{1.9cm}|p{2.8cm}|p{2.5cm}|}
        \hline
        \textbf{Dataset Name} & \textbf{Attributes} & \textbf{Data Source} \\
        \hline
        Soil Nutrition & District, Sub-district, Longitude, Latitude, Agro-ecological zone, pH, Phosphorus, Potassium, Nitrogen & Bangladesh Soil Institute \cite{web4}\\
        \hline
        Crop Nutrition & Crops, Potassium, Phosphorus, Nitrogen & Agriculture Dept. of Bangladesh \cite{web2} \\
        \hline
         Crop Production & Temperature, Rainfall, pH, Crops, Production & Agriculture Dept. of Bangladesh \\
        \hline
         Crop Disease & Region, Latitude, Longitude, Temperature, Humidity, Crop Diseases & Bangladesh Agro-Meteorological Dept. \cite{web3}\\
        \hline
         Monthly Avg. Temperature & Year, Month, Avg. Temperature & Bangladesh Meteorological Dept. \cite{web5}\\
        \hline
         Monthly Avg. Rainfall & Year, Month,
         Avg. Rainfall & Bangladesh Meteorological Dept.\\
        \hline
         Monthly Avg. Humidity & Year, Month, Avg. Humidity & Bangladesh Meteorological Dept. \\
        \hline
    \end{tabular}
    \label{tab:dataset_info}
\end{table}

\subsubsection{Meteoro-agro Region \& Soil Nutrition Dataset}
The types of soil nature and the level of different nutrient elements in the soil of all locations are not similar \cite{20}. That is why, dividing the areas into different agroecological zones based on soil nutrition level is a must to predict the production and disease possibilities for different crops in different regions.

This dataset is a compilation of information related to soil characteristics and agricultural dependencies in different regions of Bangladesh. It includes details such as the division, district, upazila (sub-district), latitude, and longitude of each location. 

Additionally, it provides information on the agroecological zone number and name, meteorological station, the pH level of the soil, and the content of phosphorus and potassium in the soil.

\subsubsection{Crop Nutrition Dataset}
The rate of production of crops depends on the amount of soil nutrition gained by crops. All crops do not require the same level of nutrition elements. Hence, we have to keep in mind the soil nutrition facts of a region to suggest the best productive crops for that region. This dataset contains information about the nutrient requirements of various crops, specifically regarding phosphorus and potassium levels. The dataset includes 18 crops along with their corresponding phosphorus and potassium levels.

\subsubsection{Crop Production Dataset}
Crop production depends on different key factors. Weather parameters and soil quality affect the production of a crop very much. So, we have created a dataset of the production of crops in ton/hector for different combinations of weather parameters and soil quality. This dataset provides information on rainfall, temperature, pH level, crop type, and corresponding production.

\subsubsection{Crop Disease Dataset}
Production of crops is greatly reduced by the impact of different crop diseases \cite{21}. If we want to build a crop-suggesting system for best production, then we must consider the possible diseases for the cultivated crop. On the other hand, diseases of crops depend on different weather parameters \cite{22}. Temperature and humidity are two of the major weather parameters affecting crop disease possibilities \cite{23}. The dataset provides specific information about different crops in different regions of Bangladesh, including the latitude and longitude coordinates, temperature, humidity, and the corresponding diseases observed under those conditions. The acquired data were not in a usable format to be used by machine learning algorithms. So, we have to create the crop disease dataset using the information of the Bangladesh Agro-meteorological Department. This dataset helps to detect the possible diseases for any crop in a given location.
\subsubsection{Monthly Average Temperature Dataset}
This dataset is composed of information about the average temperature (in \textcelsius) by month at different meteorological stations in Bangladesh. Almost 50+ years of information is placed in the dataset which has been used for future temperature prediction. The predicted temperature has been used to predict the production of crops and possible crop diseases.

\subsubsection{Monthly Average Rainfall Dataset}
This dataset is a collection of information about the average rainfall (in mm) per month at different meteorological stations in Bangladesh. There are almost 50+ years of information in this dataset for machine learning approach-based future rainfall prediction. The predicted rainfall along with the predicted temperature have been used for predicting the production of crops.

\subsubsection{Monthly Average Humidity Dataset}
Information about average the humidity (in \%) by month of different meteorological stations in Bangladesh is stated in this dataset. Like the other two weather parameter datasets, almost 50+ years of information is available in the dataset for future humidity prediction. The predicted humidity has been used to predict possible crop diseases.

\subsection{Location Based Soil Nutrition Extraction}
At first, the closest sub-district (agro-meteorological zone) to the user location is determined by applying the Haversine formula \cite{24}. If the latitude and longitude of two sites on Earth are known, the distance between them can be determined using the Haversine formula. The Haversine formula is derived from the Haversine law by using the sides and angles of spherical triangles. The Haversine formula is as follows:

\begin{equation}
     a = \sin^2\left(\frac{\Delta\phi}{2}\right) + \cos(\phi_1) \cdot \cos(\phi_2) \cdot \sin^2\left(\frac{\Delta\lambda}{2}\right) 
\end{equation}
\begin{equation}
    c = 2 \cdot \arctan2\left(\sqrt{a}, \sqrt{1-a}\right) 
\end{equation}
\begin{equation}
    d = R \cdot c 
\end{equation}
where:
\begin{itemize}
  \item $\Delta\phi$ is the difference in latitude between the user location and the agricultural zone,
  \item $\Delta\lambda$ is the difference in longitude between the user location and the agricultural zone,
  \item $\phi_1$ and $\phi_2$ are the latitudes of the user location and the agricultural zone respectively,
  \item $R$ is the radius of the Earth,
  \item $a$ is the square of half the chord length between the user location and the agricultural zone (also known as the haversine),
  \item $c$ is the angular distance in radians between the user location and the agricultural zone,
  \item $d$ is the distance between the user location and the agricultural zone in the same units as the Earth's radius.
\end{itemize}

Here we estimated the closest possible agricultural zone and meteorological station for the given user location based on latitude and longitude. It is necessary for the precision of soil nutrition identification and weather information accuracy. Then, using the Meteoro-agro Region \& Soil Nutrition Dataset, which contains information on soil nutrition according to the meteorological zone and agricultural zone, the information about the soil nutrition level of the nearest sub-district is found.

\subsection{Primary Crops Finding}
In this phase, the crop nutrition dataset is utilized to determine whether the found soil nutrition level satisfies the appropriate crops' needs or not for a given user area. A list of the major crops is thus discovered.

\subsection{Weather Parameter Forecast}
The suggested model automatically pulls the local temperature, rainfall, and humidity data from respective datasets when the user location is chosen. Rainfall, humidity, and temperature fluctuate from month to month. Furthermore, the time series data is one-dimensional. We have employed the SARIMAX (Seasonal Autoregressive Integrated Moving Average with Exogenous Variables) model, which can more accurately capture the seasonal pattern, as temperature, rainfall, and humidity are seasonal data\cite{25}. SARIMAX(p, d, q)(P, D, Q, s) can be used to specify the SARIMAX model where p, d, and q represent the successive order of differencing and moving average terms, respectively, in autoregressive terms. Additionally, P, D, Q, and s stand for the seasonal duration, seasonal moving average terms, seasonal order of differencing, and seasonal autoregressive terms respectively.

The best SARIMAX model parameters were found by using the Akaike Information Criterion (AIC) measurement. AIC is a statistician's tool for assessing the relative merits of several statistical models, especially in the context of model selection. It offers a trade-off between a model's goodness of fit and complexity.
In this work, the best SARIMAX model parameters for temperature, rainfall, and humidity are as follows:

\begin{itemize}
    \item \textbf{Temperature}: SARIMAX(1, 0, 0) (2, 1, 0, 12)
    \item \textbf{Rainfall}: SARIMAX(1, 0, 0) (0, 1, 1, 12)
    \item \textbf{Humidity}: SARIMAX(1, 0, 1) (1, 1, 0, 12)
\end{itemize}

\subsection{Crop Production Prediction}
In previous subsections, we have retrieved information on the primary cultivable crops, temperature, and rainfall. Our primary prediction model was developed at this stage. Our suggested model uses the supervised regression machine learning technique because this is the prediction of a continuous variable. Applying the Decision Tree Regression (DTR) technique, the prediction model is trained to forecast the main cultivable crops' production.

\subsection{Crop Disease Prediction}
Crops are vulnerable to diseases which vary based on weather conditions. The model is given the agricultural location along with its predicted temperature, humidity, and selected crop list based on crop and soil nutrition. The model then predicts the possible diseases and total disease count for each of the primary selected crops. We have used the Support Vector Classifier (SVC) to predict the possible disease for a list of crops.

\section{Result and Discussion}This section is focused on the discussion of the outcome of each of the steps of our work. Results acquired from each step of methodology and comparative analysis of the prediction models are shown in the following subsections.

\subsection{Experimental Results}
The suggested model was developed using Bangladesh's geography and put into practice within the setting of Bangladeshi agriculture. With the datasets we have gathered, we have assessed the model. The subsections that follow talk about the experimental results we achieved.

\subsubsection{Location Based Soil Nutrition Extraction}
Our methodology begins by finding the longitude and latitude of the user's current position. The closest sub-district is then determined by the model using the Haversine formula. Consider utilizing the Rangpur Sadar Upazila as the test location (Longitude: 25.740580 \textdegree N, Latitude: 89.261139 \textdegree E). The model selects the closest agricultural zone and retrieves the pertinent information about soil nutrition as shown in Table \ref{tab:district-info}.

\subsubsection{Primary Crops List}
The crop nutrition dataset is used to choose the main cultivable crops based on the information about soil nutrition from Table \ref{tab:district-info}. A crop will be chosen as the major cultivable crop if its nutritional value satisfies the nutritional standard for that area. The selected crops are displayed in Table \ref{tab:crop-order}.

\subsubsection{Weather Parameter Forecast}
Crop productivity and disease susceptibility are regulated by three key factors including temperature, rainfall, and humidity. SARIMAX model is used for forecasting Temperature (in \textcelsius), rainfall (in millimetres), and humidity (in \%) for the given user location which is Rangpur Sadar Upazila.  These weather parameters have been derived from three weather datasets.

\begin{table}[h]
    \centering
    \caption{EXTRACTED SOIL NUTRITION DATA}
    \begin{tabular}{|c|c|}
    \hline
    \textbf{Parameter} & \textbf{Value} \\
    \hline
    Agro-Meteoro Zone & Rangpur \\
    \hline
    Latitude & 25.740580°N \\
    \hline
    Longitude & 89.261139°E \\
    \hline
    pH & 5.6-6.5 \\
    \hline
    Phosphorus & VH \\
    \hline
    Potassium & M \\
    \hline
    \end{tabular}
    \label{tab:district-info}
\end{table}

\begin{table}[h]
    \centering
    \caption{PRIMARILY SELECTED CROPS}
    \begin{tabular}{|c|c|}
    \hline
    \textbf{Initial Order} & \textbf{Crop Name} \\
    \hline
    1 & Garlic \\
    2 & Lentil \\
    3 & Papaya \\
    4 & Rice \\
    5 & Soyabean \\
    6 & Sugarcane \\
    7 & Tomato \\
    \hline
    \end{tabular}
    \label{tab:crop-order}
\end{table}

\begin{table}[h]
    \centering
    \caption{EXTRACTED WEATHER DATA}
    \begin{tabular}{|c|c|c|c|}
    \hline
    \textbf{Month} & \textbf{Temp.(\textdegree C)} & \textbf{Rain.(mm)} & \textbf{Hum.(\%)} \\
    \hline
    Jan & 15.8 & 0 & 82 \\
    \hline
    Feb & 20.5 & 10 & 75 \\
    \hline
    Mar & 23.7 & 24 & 68 \\
    \hline
    Apr & 26.6 & 94 & 77 \\
    \hline
    May & 27.4 & 232 & 82 \\
    \hline
    Jun & 29 & 289 & 80 \\
    \hline
    Jul & 28.4 & 542 & 83 \\
    \hline
    Aug & 28.4 & 572 & 85 \\
    \hline
    Sep & 28 & 299 & 84 \\
    \hline
    Oct & 27.1 & 116 & 84 \\
    \hline
    Nov & 22.6 & 3 & 78 \\
    \hline
    Dec & 18.4 & 0 & 80 \\
    \hline
    \end{tabular}
    \label{tab:climate-data}
\end{table}

The temperature, rainfall, and humidity statistics of the user location for 2023 are displayed in Table \ref{tab:climate-data}.

\subsubsection{Crop Disease Prediction}
By using the temperature and humidity data for user location from Table \ref{tab:climate-data} and primary selected crops list from Table \ref{tab:crop-order}, the possible diseases for each crop are suggested in Table \ref{tab:crop-count-disease}.

\subsubsection{Crop Production Prediction}
The soil nutrition, primary cultivable crops, expected temperature, and rainfall are all obtained from the aforementioned procedures. In this stage, the system trains the proposed regression model to forecast the production amount of the primary selected crops by using the crop production dataset.

\subsubsection{Final Suggested Crops List}
The model then arranges the predicted production for different crops in decreasing order. The model generates the final list of crops displayed in Table \ref{tab:crop-production} for the stated user location (Rangpur Sadar Upazila).

\begin{table}[h]
    \centering
    \caption{PREDICTED CROP DISEASES }
    \begin{tabular}{|c|c|c|c|}
    \hline
    \textbf{Initial Order} & \textbf{Crop} & \textbf{Count} & \textbf{Disease} \\
    \hline
    1 & Garlic & 0 & - \\
    2 & Lentil & 1 & Foot rot \\
    3 & Papaya & 0 & - \\
    4 & Rice & 0 & - \\
    5 & Soyabean & 1 & Anthracnose \\
    6 & Sugarcane & 1 & Smut \\
    7 & Tomato & 0 & - \\
    \hline
    \end{tabular}
    
    \label{tab:crop-count-disease}
\end{table}

\begin{table}[h]
    \centering
    \caption{FINAL CROPS LIST }
    \begin{tabular}{|c|c|c|c|}
    \hline
    \textbf{Final Order} & \textbf{Crop} & \textbf{Production (ton/hectare)} & \textbf{Disease} \\
    \hline
    1 & Papaya & 134.24 & Not found \\
    2 & Sugarcane & 106.79 & Smut \\
    3 & Tomato & 35.17 & Not found \\
    4 & Garlic & 12.79 & Not found \\
    5 & Soyabean & 11.44 & Anthracnose \\
    6 & Rice & 7.99 & Not found \\
    7 & Lentil & 0.85 & Foot rot \\
    \hline
    \end{tabular}
    \label{tab:crop-production}
\end{table}

\subsection{Comparative Analysis of Disease Prediction Models}
Different classification models have been applied to find the best-suited model for predicting the possible diseases of the suggested crops. A few well-known factors have been considered for this purpose which are stated below:
\begin{itemize}
\item \textbf{Accuracy}: A classification model's overall correctness can be estimated by its accuracy. It determines the proportion of accurately predicted occurrences to all of the dataset's instances.
\item \textbf{Precision}: Precision is the measurement of the accuracy of positive predictions given by the model. It is calculated as the proportion of true positives to the total number of positive predictions (true positives + false positives) from the whole dataset.
\item \textbf{Recall}: Recall represents the ability of a model to accurately recognize all relevant instances in the dataset. It can be measured as the ratio of true positives to the total number of real positives (true positives + false negatives).
\item \textbf{F1 Score}: The harmonic mean of precision and recall can be stated as the F1 score. The trade-off between precision and recall is balanced by this F1 score and it suggests a single summarizing score about a model's performance.
\end{itemize}

Using the crop disease dataset, the Support Vector Classifier (SVC), Random Forest Classifier (RFC), Gradient Boosting Classifier (GBC), and Logistic Regression (LoR) model are trained and tested on the evaluation metrics.  The findings of the comparative error analysis are shown in Table \ref{tab:model-evaluation-disease}.  

Most often, higher values of accuracy, precision, recall, and F1 score indicate the best classification model. For this work, SVC provides the best disease predictions rather than other models.

\subsection{Comparative Analysis of Production Prediction Models}
To forecast the production of the primarily selected crops, different regression methods have been compared to choose the most compatible model. The most common method of assessment is to estimate how well the projected value matches the actual value. The following metrics for evaluation are investigated for choosing the best model:

\begin{itemize}
\item \textbf{Mean Squared Error (MSE)}: The mean square difference between predicted and actual values for regression algorithms is known as MSE.
\item \textbf{Root Mean Squared Error(RMSE)}: This error type provides the root mean
square difference between the anticipated and real values.
\item \textbf{R-squared}: R-squared value shows how appropriate the data is to the fitted regression line.
\end{itemize}

\begin{table}[h]
    \centering
    \caption{EVALUATION METRICS FOR THE DISEASE PREDICTION MODEL}
    \begin{tabular}{|c|c|c|c|c|}
    \hline
    \textbf{Model} &\textbf{Accuracy} & \textbf{Precision} & \textbf{Recall} & \textbf{F1 Score} \\
    \hline
    SVC & 0.94 & 0.91 & 0.94 & 0.92 \\
    RFC & 0.91 & 0.89 & 0.91 & 0.90 \\
    GBC & 0.90 & 0.89 & 0.90 & 0.90 \\
    LoR & 0.56 & 0.59 & 0.56 & 0.57 \\
    \hline
    \end{tabular}
    \label{tab:model-evaluation-disease}
\end{table}

\begin{table}[h]
    \centering
    \caption{EVALUATION METRICS FOR THE PRODUCTION PREDICTION MODEL}
    \begin{tabular}{|c|c|c|c|}
    \hline
    \textbf{Model} & \textbf{MSE} & \textbf{RMSE} & \textbf{R-Squared} \\
    \hline
    DTR & 1.06 & 1.03 & 0.997 \\
    RFR & 1.19 & 1.09 & 0.996 \\
    LR & 42.89 & 6.54 & 0.891 \\
    GBR & 4.54 & 2.13 & 0.988 \\
    \hline
    \end{tabular}
    \label{tab:model-evaluation}
\end{table}

The optimal regression procedure for the final forecast is what we are aiming for. Using the crop production dataset, the Linear Regression (LR), Decision Tree Regression (DTR), Gradient Boosting Regression (GBR), and Random Forest Regression (RFR) models are trained and tested. The findings of the comparative error analysis are shown in Table \ref{tab:model-evaluation}. 

The lower MSE and RMSE values suggest that the forecasted result and the real outcome are quite close to each other. The R-squared value, on the other hand, ranges from 0 to 1. The greater the number is, the more accurately the model is fitted. DTR is the ideal model for our system, according to the findings.

\section{Conclusion}
In this work, we have created a model to recommend the most productive crops along with probable crop diseases based on the user's location. Bangladesh serves as the backdrop for the creation of the whole work. The fast-expanding population increases food consumption, yet the amount of arable land is steadily shrinking in this country.
Therefore, increasing agricultural output is now unavoidable. In this work, a model that suggests economically and more productive crops throughout the year in any place has been developed using machine learning techniques. 

We utilized SARIMAX as the rainfall, humidity, and temperature forecasting model for the weather forecast. Then, probable disease is predicted based on temperature and humidity for the primary selected crops. At last, crop production is determined using a regression model based on soil nutrition data and forecasted temperature and rainfall data. Compared to other models, the SVC model shows better outcomes for crop disease prediction and the DTR model performs better in predicting crop yield. Hopefully, the application of our strategy to the agriculture sector would increase crop yield and result in economic growth. We intend to provide Bangladeshi farmers access to this technology through a mobile platform in the future. Converting this work into the form of software will be more beneficial and easier for the farmers to know which crops they should cultivate to maximize their economic profit and also get prepared for possible crop diseases. Besides, the work can be extended by using more crop and disease data along with more weather parameter dependency.

\bibliographystyle{IEEEtran}
\bibliography{References}

\end{document}